\documentclass[10pt,twocolumn,letterpaper]{article}

\usepackage{cvpr}
\usepackage{times}
\usepackage{epsfig}
\usepackage{graphicx}
\usepackage{amsmath}
\usepackage{amssymb}
\usepackage{subfigure}

\usepackage{caption}

\usepackage{epsfig}
\usepackage{graphicx}
\usepackage{algorithm}
\usepackage{algorithmic}

\def\E{{\rm E}}

\usepackage[pagebackref=true,breaklinks=true,letterpaper=true,colorlinks,bookmarks=false]{hyperref}

\cvprfinalcopy % *** Uncomment this line for the final submission

\def\cvprPaperID{5829} % *** Enter the CVPR Paper ID here

\ifcvprfinal\fi
\begin{document}
%\captionsetup[subfigure]{labelformat=empty}
%%%%%%%%% TITLE
\title{Joint Training of Variational Auto-Encoder and Latent Energy-Based Model}

\author{Tian Han$^1$, Erik Nijkamp$^2$,\, Linqi Zhou$^2$, Bo Pang$^2$,\,  Song-Chun Zhu$^2$, Ying Nian Wu$^2$\\
%Department of Statistics\\
$^1$Department of Computer Science, Stevens Institute of Technology \\
$^2$Department of Statistics, University of California, Los Angeles\\
{\tt\small than6@stevens.edu, \{enijkamp,bopang\}@ucla.edu, linqizhou907@gmail.com} \\
{\tt\small\{sczhu,ywu\}@stat.ucla.edu}
}

\maketitle
%\thispagestyle{empty}

%%%%%%%%% ABSTRACT
\begin{abstract}

This paper proposes a joint training method to learn both the variational auto-encoder (VAE) and the latent energy-based model (EBM). %The VAE consists of a generator network, which transforms a latent noise vector to an image, as well as an inference model that infers the latent vector from the image. The latent EBM defines an unnormalized joint distribution on the latent vector and the image based on a joint energy function. 
The joint training of VAE and latent EBM are based on an objective function that consists of three Kullback-Leibler divergences between three joint distributions on the latent vector and the image, and the objective function is of an elegant symmetric and anti-symmetric form of divergence triangle that seamlessly integrates variational and adversarial learning. In this joint training scheme, the latent EBM serves as a critic of the generator model, while the generator model and the inference model in VAE serve as the approximate synthesis sampler and inference sampler of the latent EBM. Our experiments show that the joint training greatly improves the synthesis quality of the VAE. It also enables learning of an energy function that is capable of detecting out of sample examples for anomaly detection. 
   
\end{abstract}

%%%%%%%%% BODY TEXT
\section{Introduction}

The variational auto-encoder (VAE)~\cite{kingma2013auto,rezende2014stochastic} is a powerful method for generative modeling and unsupervised learning. It consists of a generator model that transforms a noise vector to a signal such as image via a top-down convolutional network (also called deconvoluitional network due to its top-down nature). It also consists of an inference model that infers the latent vector from the image via a bottom-up network. %The inference network has been generalized to flow-based model for improved inference. 
The VAE has seen many applications in image and video synthesis~\cite{gregor2015draw,babaeizadeh2017stochastic} and unsupervised and semi-supervised learning~\cite{sonderby2016ladder,kingma2014semi}. 

Despite its success, the VAE suffers from relatively weak synthesis quality compared to methods such as generative adversarial net (GANs)~\cite{goodfellow2014generative,radford2015unsupervised} that are based on adversarial learning. While combing VAE objective function with the GAN objective function can improve the synthesis quality, such a combination is rather ad hoc. In this paper, we shall pursue a more systematic integration of variational learning and adversarial learning. Specifically, instead of employing a discriminator as in GANs, we recruit a latent energy-based model (EBM) to mesh with VAE seamlessly in a joint training scheme. 

The generator model in VAE is a directed model, with a known prior distribution on the latent vector, such as Gaussian white noise distribution, and a conditional distribution of the image given the latent vector. The advantage of such a model is that it can generate synthesized examples by direct ancestral sampling. The generator model defines a joint probability density of the latent vector and the image in a top-down scheme. We may call this joint density the generator density. 

VAE also has an inference model, which defines the conditional distribution of the latent vector given the image. Together with the data distribution that generates the observed images, they define a joint probability density of the latent vector and the image in a bottom-up scheme. We may call this joint density the joint data density, where the latent vector may be considered as the missing data in the language of the EM algorithm~\cite{dempster1977maximum}. 

As we shall explain later, the VAE amounts to joint minimization of the Kullback-Leibler divergence from data density to the generator density, where the joint minimization is over the parameters of both the generator model and the inference model. In this minimization, the generator density seeks to cover the modes of the data density, and as a result, the generator density can be overly dispersed. This may partially explain VAE's lack of synthesis quality. 

Unlike the generator network, the latent EBM is an undirected model. It defines an unnormalized joint density on the latent vector and the image via a joint energy function. Such undirected form enables the latent EBM to better approximate the data density than the generator network. However, the maximum likelihood learning of latent EBM requires (1) inference sampling: sampling from the conditional density of the latent vector given the observed example and (2) synthesis sampling: sampling from the joint density of the latent vector and the image. Both inference sampling and synthesis sampling require time consuming Markov chain Monte Carlo (MCMC). 

In this paper, we propose to jointly train the VAE and the latent EBM so that these two models can borrow strength from each other. The objective function of the joint training method consists of the Kullback-Leibler divergences between three joint densities of the latent vector and the image, namely the data density, the generator density, and the latent EBM density. The three Kullback-Leilber divergences form an elegant symmetric and anti-symmetric form of divergence triangle that integrates the variational learning and the adversarial learning seamlessly. 

The joint training is beneficial to both the VAE and the latent EBM. The latent EBM has a more flexible form and can approximate the data density better than the generator model. It serves as a critic of the generator model by judging it against the data density. To the generator model, the latent EBM serves as a surrogate of the data density and a target density for the generator model to approximate. The generator model and the associated inference model, in return, serve as approximate synthesis sampler and inference sampler of the latent EBM, thus relieving the latent EBM of the burden of MCMC sampling. 

Our experiments show that the joint training method can learn the generator model with strong synthesis ability. It can also learn energy function that is capable of anomaly detection. 

\section{Contributions and related work} 

The following are contributions of our work. (1) We propose a joint training method to learn both VAE and latent EBM.  The objective function is of a symmetric and anti-symmetric form of divergence triangle. (2) The proposed method integrates variational and adversarial learning. (3) The proposed method integrates the research themes initiated by the Boltzmann machine and Helmholtz machine. 

The following are the themes that are related to our work. 

(1) Variational and adversarial learning. Over the past few years, there has been an explosion in research on variational learning and adversarial learning, inspired by VAE~\cite{kingma2013auto,rezende2014stochastic,sonderby2016ladder,gregor2015draw} and GAN~\cite{goodfellow2014generative,radford2015unsupervised,arjovsky2017wasserstein,zhao2016energy} respectively. One aim of our work is to find a natural integration of variational and adversarial learning. We also compare with some prominent methods along this theme in our experiments. Notably, adversarially learned inference (ALI)~\cite{dumoulin2016adversarially,donahue2016adversarial} combines the learning of the generator model and inference model in an adversarial framework. It can be improved by adding conditional entropy regularization as in more recent methods ALICE~\cite{li2017alice} and SVAE~\cite{chen2018symmetric}. Though these methods are trained using joint discriminator on image and latent vector, such a discriminator is not a probability density, thus it is not a latent EBM. 

(2) Helmholtz machine and Boltzmann machine.   Before VAE and GAN took over, Boltzmann machine~\cite{ackley1985learning,hinton2006fast,salakhutdinov2010efficient} and Helmholtz machine~\cite{hinton1995wake} are two classes of models for generative modeling and unsupervised learning. Boltzmann machine is the most prominent example of latent EBM. The learning consists of two phases. The positive phase samples from the conditional distribution of the latent variables given the observed example. The negative phase samples from the joint distribution of the latent variables and the image. The parameters are updated based on the differences between statistical properties of positive and negative phases. Helmholtz machine can be considered a precursor of VAE. It consists of a top-down generation model and the bottom-up recognition model. The learning also consists of two phases. The wake phase infers the latent variables based on the recognition model and updates the parameters of the generation model. The sleep phase generates synthesized data from the generation model and updates the parameters of the recognition model. Our work seeks to integrate the two classes of models. 

(3) Divergence triangle for joint training of generator network and energy-based model. The generator and energy-based model can be trained separately using maximum likelihood criteria as in~\cite{han2017alternating,nijkamp2019anatomy}, and they can also be trained jointly as recently explored by ~\cite{kim2016deep,xie2016cooperative,han2019divergence,kumar2019maximum}. In particular, \cite{han2019divergence} proposes a divergence triangle criterion for joint training. Our training criterion is also in the form of divergence triangle. However, the EBM in these papers is only defined on the image and there is no latent vector in the EBM. In our work, we employ latent EBM which defines a joint density on the latent vector and the image, thus this undirected joint density is more natural match to the generator density and the data density, both are directed joint densities of the latent vector and the image. 

%The joint training of generator network and energy-based mode has been explored recently by
\section{Models and learning}

\subsection{Generator network} 
\label{sec:generator}
Let $z$ be the $d$-dimensional latent vector. Let $x$ be the $D$-dimensional signal, such as an image. VAE consists of a generator model, which defines a joint probability density 
\begin{align} 
   p_\theta(x, z) = p(z) p_\theta(x|z), 
\end{align} 
in a top-down direction, where $p(z)$ is the known prior distribution of the latent vector $z$, such as uniform distribution or Gaussian white noise, i.e., $z \sim {\rm N}(0, I_d)$, where $I_d$ is the $d$-dimensional identity matrix. $p_\theta(x|z)$ is the conditional distribution of $x$ given $z$. A typical form of $p_\theta(x|z)$ is such that 
$
   x = g_\theta(z) + \epsilon, 
$
where $g_\theta(z)$ is parametrized by a top-down convolutional network (also called the deconvolutional network due to the top-down direction), with $\theta$ collecting all the weight and bias terms of the network. $\epsilon$ is the residual noise, and usually it is assumed $\epsilon \sim {\rm N}(0, \sigma^2 I_D)$. 

The generator network is a directed model. We call $p_\theta(x, z)$ the generator density. $x$ can be sampled directly by first sampling $z$ and then sampling $x$ given $z$. This is sometimes called ancestral sampling in the literature~\cite{mohamed2016learning}. 

The marginal distribution $p_\theta(x) = \int p_\theta(x, z) dz$. It is not in closed form. Thus the generator network is sometimes called the implicit generative model in the literature. 

The inference of $z$ can be based on the posterior distribution of $z$ given $x$, i.e., $p_\theta(z|x) = p_\theta(x, z)/p_\theta(x)$. $p_\theta(z|x)$ is not in closed form. %Sampling $p_\theta(z|x)$ usually requires MCMC. 

\subsection{Inference model} 
\label{sec:inference}
VAE assumes an inference model $q_\phi(z|x)$ with a separate set of parameters $\phi$. One example of $q_\phi(z|x)$ is ${\rm N}(\mu_\phi(x), V_\phi(x))$, where $\mu_\phi(x)$ is the $d$-dimensional mean vector, and $V_\phi(x)$ is the $d$-dimensional diagonal variance-covariance matrix. Both $\mu_\phi$ and $V_\phi$ can be parametrized by bottom-up convolutional networks, whose parameters are denoted by $\phi$. The inference model $q_\phi(z|x)$ is a closed form approximation to the true posterior $p_\theta(z|x)$. %To improve the approximation, flow-based models have been added on top of the above multivariate normal distribution~\cite{rezende2015variational,dinh2016density}. 

\subsection{Data density} 

Let $q_{\rm data}(x)$ be the distribution that generates the observed images. In practice, expectation with respect to $q_{\rm data}(x)$ can be approximated by the average over the observed training examples. 

The reason we use the notation $q$ to denote the data distribution $q_{\rm data}(x)$ is that $q_{\rm data}(x)$ can be naturally combined with the inference model $q_\phi(z|x)$, so that we have the joint density 
\begin{align} 
    q_\phi(x, z) = q_{\rm data}(x) q_\phi(z|x). 
\end{align}
The above is also a directional density in that it can be factorized in a bottom-up direction. We may call the joint density $q_\phi(x, z)$ the data density, where in the terminology of the EM algorithm, we may consider $z$ as the missing data, and $q_\phi(z|x)$ as the imputation model of the missing data. 

\subsection{VAE} 

The top-down generator density $p_\theta(x, z) = p(z) p_\theta(x|z)$ and the bottom-up data density $q_\phi(x, z) = q_{\rm data}(x) q_\phi(z|x)$ form a natural pair. As noted by~\cite{han2019divergence}, VAE can be viewed as the following joint minimization 
\begin{align} 
  \min_\theta \min_\phi {\rm KL}(q_\phi(x, z) \| p_\theta(x, z)), 
\end{align} 
where for two densities $q(x)$ and $p(x)$ in general, ${\rm KL}(q(x)\|p(x)) = \E_q[\log (q(x)/p(x)]$ is the Kullback-Leibler divergence between $q$ and $p$.% (or from $q$ to $p$ since ${\rm KL}(q\|p)$ is asymmetric). 

To connect the above joint minimization to the usual form of VAE, 
\begin{align} 
  {\rm KL}(q_\phi(x, z) \| &p_\theta(x, z)) = {\rm KL}(q_{\rm data}(x) \| p_\theta(x)) \\
  &+  \E_{q_{\rm data}(x)}[ {\rm KL}(q_{\phi}(z|x) \| p_\theta(z|x)) ],
\end{align}
where for two joint densities $q(x, y)$ and $p(x, y)$, we define ${\rm KL}(q(y|x)\| p(y|x)) = \E_{q(x|y)} [\log (q(y|x)/p(y|x)) \geq 0$. % (some textbooks~\cite{cover2012elements} define ${\rm KL}(q(y|x)\| p(y|x)) = \E_{q(x, y)} [\log (q(y|x)/p(y|x))$, where the expectation is with respect to $\E_{q(x, y)}= \E_{q(x)} \E_{q(y|x)}$. We feel our definition is more clear even though it is a bit tedious). 

Since ${\rm KL}(q_{\rm data}(x)\|p_\theta(x)) = \E_{q_{\rm data}(x)}[\log q_{\rm data}(x)] - \E_{q_{\rm data}(x)}[\log p_\theta(x)]$, the joint minimization problem is equivalent to the joint maximization of 
\begin{align} 
   & \E_{q_{\rm data}(x)}[\log p_\theta(x) - {\rm KL}(q_{\phi}(z|x) \| p_\theta(z|x))] \\
   & = \E_{q_{\rm data}(x)}[ \E_{q_\phi(z|x)} [ \log p_\theta(x, z)] -  \E_{q_\phi(z|x)}[ \log q_\phi(z|x)], \nonumber
\end{align}
which is the lower bound of the log-likelihood used in VAE~\cite{kingma2013auto}. %The maximization of $\phi$ given $\theta$ is to minimize $\E_{q_{\rm data}(x)}[{\rm KL}(q_\phi(z|x)) \| p_\theta(z|x))]$. The maximization of $\theta$ given $\phi$ is to maximize $\E_{q_{\rm data}(x) q_\phi(z|x)}[\log p_\theta(x, z)]$. 

%The above joint minimization can be interpreted as alternating projection. Given $\phi$, the minimization of $\theta$ is to fit the generator model based on the imputed or inferred $z \sim q_\phi(z|x)$. This step corresponds to the M-step of EM. Given $\theta$, the minimization of $\phi$ is to find the inference or imputation model $q_\phi(z|x)$. This step corresponds to the E-step of EM. 
It is worth of noting that the wake-sleep algorithm ~\cite{hinton1995wake} for training the Helmholtz machine consists of (1) wake phase:   $\min_\theta {\rm KL}(q_\phi(x, z) \| p_\theta(x, z))$, and (2) sleep phase: $\min_\phi {\rm KL}(p_\theta(x, z) \|q_\phi(x, z))$. The sleep phase reverses the order of KL-divergence.

\subsection{Latent EBM} 

Unliked the directed joint densities $p_\theta(x, z) = p(z) p_\theta(x|z)$ in the generator network, and $q_\phi(x, z) = q_{\rm data}(x) q_\phi(z|x)$ in the data density, the latent EBM defines an undirected joint density, albeit an unnormalized one: 
\begin{align} 
  \pi_\alpha(x, z) = \frac{1}{Z(\alpha)} \exp[ f_\alpha(x, z) ], 
\end{align} 
where $- f_\alpha(x, z)$ is the energy function (a term originated from statistical physics) defined on the image $x$ and the latent vector $z$. $Z(\alpha) = \int \int \exp[f_\alpha(x, z)] dx dz$ is the normalizing constant. It is usually intractable, and $\exp[f_\alpha(x, z)]$ is an unnormalized density. The most prominent example of latent EBM is the Boltzmann machine~\cite{ackley1985learning,hinton2006fast}, where $f_\alpha(x, z)$ consists of pairwise potentials. In our work, we first encode $x$ into a vector and then concatenate this vector with the vector $z$, and then get $f_\alpha(x, z)$ by a network defined on the concatenated vector. 

\subsection{Inference and synthesis sampling} 

Let $\pi_\alpha(x) = \int \pi_\alpha(x, z) dz$ be the marginal density of latent EBM.  The maximum likelihood learning of $\alpha$ is based on $\min_\alpha {\rm KL}(q_{\rm data}(x) \|\pi_\alpha(x))$ because minimizing ${\rm KL}(q_{\rm data}(x)\|\pi_\alpha(x))$ is equivalent to maximizing the log-likelihood $\E_{q_{\rm data}(x)}[\log \pi_\alpha(x)]$. The learning gradient is 
\begin{align} 
  \frac{\partial}{\partial \alpha} \E_{q_{\rm data}(x)}[\log \pi_\alpha(x)] &= \E_{q_{\rm data}(x) \pi_\alpha(z|x)} \left[   \frac{\partial}{\partial \alpha} f_\alpha(x, z) \right] \nonumber \\
  &-\E_{\pi_\alpha(x, z)} \left[   \frac{\partial}{\partial \alpha} f_\alpha(x, z) \right].\label{eq:L}
  \end{align}
This is a well known result in latent EBM~\cite{ackley1985learning,lecun2006tutorial}. %We give a derivation in the supplementary material. 
On the right hand of the above equation, the two expectations can be approximated by Monte Carlo sampling. For each observed image, sampling from $\pi_\alpha(z|x)$ is to infer $z$ from $x$. We call it the inference sampling. In the literature,  it is called the positive phase ~\cite{ackley1985learning,hinton2006fast}. It is also called clamped sampling where $x$ is an observed image and is fixed. Sampling from $\pi_\alpha(x, z)$ is to generate synthesized examples from the model. We call it the synthesis sampling. In the literature,  it is called the negative phase. It is also called unclamped sampling where $x$ is also generated from the model.

\section{Joint training} 

\subsection{Objective function of joint training} 

We have the following three joint densities. 

(1) The generator density $p_\theta(x, z) = p(z) p_\theta(x|z)$. 

(2) The data density $q_\phi(x, z) = q_{\rm data}(x) q_\phi(z|x)$. 

(3) The latent EBM density $\pi_\alpha(x, z)$. 

We propose to learn the generator model parametrized by $\theta$, the inference model parametrized by $\phi$,  and the latent EBM parametrized by $\alpha$ by the following divergence triangle:   
\begin{align} 
    & \min_\theta  \min_\phi \max_\alpha L(\theta, \alpha, \phi), \\
    & L = {\rm KL}(q_\phi\|p_\theta) + {\rm KL}(p_\theta \| \pi_\alpha) -  {\rm KL}(q_\phi \| \pi_\alpha),
\end{align} 
where all the densities $q_\phi$, $p_\theta$, and $\pi_\alpha$ are joint densities of $(x, z)$. 

The above objective function is in an symmetric and anti-symmetric form. The anti-symmetry is caused by the negative sign in front of ${\rm KL}(q_\phi \| \pi_\alpha)$ and the maximization over $\alpha$. 

\subsection{Learning of latent EBM} 

For learning the latent EBM, $\max_\alpha L$ is equivalent to minimizing 
\begin{align} 
 L_{\rm E}(\alpha) = {\rm KL}(q_\phi \| \pi_\alpha) - {\rm KL}(p_\theta \| \pi_\alpha).
\end{align}
In the above minimization, $\pi_\alpha$ seeks to get close to the data density $q_\phi$ and get away from $p_\theta$. Thus $\pi_\alpha$ serves as a critic of $p_\theta$ by comparing $p_\theta$ against $q_\phi$. Because of the undirected form of $\pi_\alpha$, it can be more flexible than the directional $p_\theta$ in approximating $q_\phi$. 

The gradient of the above minimization is 
\begin{align} 
 \frac{d}{d \alpha} L_{\rm E}(\alpha) &= -\E_{q_{\rm data}(x) q_\phi(z|x)} \left[   \frac{\partial}{\partial \alpha} f_\alpha(x, z) \right] \nonumber \\
  &+\E_{p_\theta(x, z)} \left[   \frac{\partial}{\partial \alpha} f_\alpha(x, z) \right]. \label{eq:A}
\end{align} 
%Please see supplementary material for a derivation. 

Comparing Eqn.\ref{eq:A} to Eqn.\ref{eq:L}, we replace $\pi_\alpha(z|x)$ by $q_\phi(z|x)$ in the inference sampling in the positive phase, and we replace $\pi_\alpha(x, z)$ by $p_\theta(x, z)$ in the synthesis sampling in the negative phase. Both $q_\phi(z|x)$ and $p_\theta(x, z)$ can be sampled directly. Thus the joint training enables the latent EBM to avoid MCMC in both inference sampling and synthesis sampling. In other words, the inference model serves as an approximate inference sampler for latent EBM, and the generator network serves as an approximate synthesis sampler for latent EBM. 

\subsection{Learning of generator network} 

%As explained before, $\min_\theta \min_\phi {\rm KL}(q_\phi\|p_\theta)$ leads to variational learning of VAE. That is, $q_\phi$ and $p_\theta$ move toward each other. 

For learning the generator network, $\min_\theta L$ is equivalent to minimizing 
\begin{align}
\label{eq:g} 
   L_{\rm G}(\theta) = {\rm KL}(q_\phi \| p_\theta) + {\rm KL}(p_\theta \| \pi_\alpha). 
\end{align} 
where the gradient can be computed as:
\begin{align}
\label{eq:grad_g}
 \frac{d}{d \theta} L_{\rm G}(\theta) &= -\E_{q_{\rm data}(x) q_\phi(z|x)}\left[ \frac{\partial}{\partial \theta}\log p_\theta(x|z) \right] \nonumber \\
 &- \frac{\partial}{\partial \theta}\E_{p_\theta(x, z)}\left[ f_\alpha(x, z) \right].
\end{align}

 In ${\rm KL}(q_\phi \| p_\theta)$, $p_\theta$ appears on the right hand side of KL-divergence. Minimizing this KL-divergence with respect to $\theta$ requires $p_\theta$ to cover all the major modes of $q_\phi$. If $p_\theta$ is not flexible enough, it will strain itself to cover all the major modes, and as a result, it will make $p_\theta$ over-dispersed than $q_\phi$. This may be the reason that VAE tends to suffer in synthesis quality. 
 
 However, in the second term, ${\rm KL}(p_\theta \| \pi_\alpha)$, $p_\theta$ appears on the left hand side of KL-divergence, and $\pi_\alpha$, which seeks to get close to $q_\phi$ and get away from $p_\theta$ in its dynamics, serves as a surrogate for data density $q_\phi$, and a target for $p_\theta$. Because $p_\theta$ appears on the left hand side of KL-divergence, it has the mode chasing behavior, i.e., it may chase some major modes of $\pi_\alpha$ (surrogate of $q_\phi$), while it does not need to cover all the modes. Also note that in ${\rm KL}(p_\theta \| \pi_\alpha)$, we do not need to know $Z(\alpha)$ because it is a constant as far as $\theta$ is concerned. 
 
 Combine the above two KL-divergences, approximately, we minimize a symmetrized version of KL-divergence $S(q_\phi \| p_\theta) = {\rm KL}(q_\phi \| p_\theta) + {\rm KL}(p_\theta\| q_\phi)$ (assuming $\pi_\alpha$ is close to $q_\phi$). This will correct the over-dispersion of VAE, and improve the synthesis quality of VAE. 
 
 We refer the reader to the textbook~\cite{goodfellow2016deep,murphy2012machine} on the difference between $\min_p {\rm KL}(q\|p)$ and $\min_p {\rm KL}(p\|q)$. In the literature, they are also referred to as inclusive and exclusive KL, or KL and reverse KL. 
 
 \subsection{An adversarial chasing game} 
 
 The dynamics of $\pi_\alpha$ is that it seeks to get close to the data density $q_\phi$ and get away from $p_\theta$. But the dynamics of $p_\theta$ is that it seeks to get close to $\pi_\alpha$ (and at the same time also get close to the data density $q_\phi$). This defines an adversarial chasing game, i.e., $\pi_\alpha$ runs toward $q_\phi$ and runs from $p_\theta$, while $p_\theta$ chases $\pi_\alpha$. As a result, $\pi_\alpha$ leads $p_\theta$ toward $q_\phi$. $p_\theta$ and $\pi_\alpha$ form an actor-critic pair. 
 
 \subsection{Learning of inference model} 
 
 The learning of the inference model $q_\phi(z|x)$ can be based on $\min_\phi L(\theta, \alpha, \phi)$, which is equivalent to minimizing
 \begin{align} 
 \label{eq:inf}
     L_{\rm I}(\phi) =  {\rm KL}(q_\phi \|p_\theta) - {\rm KL}(q_\phi \| \pi_\alpha). 
 \end{align} 
 $q_\phi(z|x)$ seeks to be close to $p_\theta(z|x)$ relative to $\pi_\alpha(z|x)$. That is, $q_\phi(z|x)$ seeks to be the inference model for $p_\theta$. Meanwhile, $\pi_\alpha(z|x)$ seeks to be close to $q_\phi(z|x)$. This is also a chasing game. $q_\phi(z|x)$ leads $\pi_\alpha(z|x)$ to be close to $p_\theta(z|x)$. 
 
The gradient of $L_{\rm I}(\phi)$ in Eqn.\ref{eq:inf} can be readily computed as:
\begin{align}
\label{eq:grad_I}
\frac{d}{d \phi} L_{\rm I}(\phi) &= \frac{\partial}{\partial \phi}\E_{q_\phi(x, z)}\left[ \log q_{\phi}(x, z) - \log p_\theta(x, z)\right] \nonumber \\
&- \frac{\partial}{\partial \phi}\E_{q_\phi(x, z)}\left[ \log q_{\phi}(x, z) - f_\alpha(x, z)\right].
\end{align}

 We may also learn $q_\phi(z|x)$ by minimizing 
  \begin{align} 
 \label{eq:inf1}
      L_{\rm I}(\phi) =  {\rm KL}(q_\phi \|p_\theta) + {\rm KL}(q_\phi \| \pi_\alpha),  
 \end{align} 
 where we let $q_\phi(z|x)$ to be close to both $p_\theta(z|x)$ and $\pi_\alpha(z|x)$ in variational approximation.

\subsection{Algorithm} 
{\small
 \begin{algorithm}
	\caption{Joint Training for VAE and latent EBM}
	\label{alg:vae_latent}
	\begin{algorithmic}[1]
		\REQUIRE ~~\\
		training images $\{x_i\}_{i=1}^{n}$;
		number of learning iterations $T$;
		$\alpha$, $\theta$, $\phi \leftarrow$ initialized network parameters.  
		\ENSURE~~\\
		estimated parameters $\{\alpha, \theta, \phi\}$;
		generated samples $\{\tilde{x}_i\}_{i=1}^{\tilde{n}}$.	
		\STATE Let $t \leftarrow 0$.
		\REPEAT 
		\STATE {\bf Synthesis sampling for $(z_i, \tilde{x}_i)_{i=1}^{\tilde{M}}$} using Eqn.\ref{eq:sample}.
		
		%$\{z_i \sim p(z)\}_{i=1}^{\tilde{M}}$, $\{\tilde{x}_i \sim p_\theta(x|z_i)\}_{i=1}^{\tilde{M}}$.
		\STATE {\bf Inference sampling for $(x_i, \tilde{z}_i)_{i=1}^{M}$} using Eqn.\ref{eq:sample}.
		
		%$\{x_i \sim q_{\rm data}(x)\}_{i=1}^{M}$, $\{\tilde{z}_i \sim q_\phi(z|x_i)\}_{i=1}^{M}$.
		
		\STATE {\bf Learn latent EBM}: Given $\{z_i, \tilde{x}_i\}_{i=1}^{\tilde{M}}$ and $\{x_i, \tilde{z}_i\}_{i=1}^{M}$, update $\alpha \leftarrow \alpha - \eta_\alpha L'_{\rm E}(\alpha)$ using Eqn.~\ref{eq:A_approx} with learning rate $\eta_\alpha$. 
		\STATE {\bf Learn inference model}:  Given $\{x_i, \tilde{z}_i\}_{i=1}^{M}$, 
				update $\phi \leftarrow \phi - \eta_\phi  L'_{\rm I}(\phi)$, with learning rate $\eta_\phi$ using Eqn.~\ref{eq:grad_I_approx}.
		\STATE {\bf Learn generator network}: Given $\{z_i, \tilde{x}_i\}_{i=1}^{\tilde{M}}$ and $\{x_i, \tilde{z}_i\}_{i=1}^{M}$, 
				update $\theta \leftarrow \theta - \eta_\theta L'_{\rm G}(\theta)$, with learning rate $\eta_\theta$ using Eqn.~\ref{eq:grad_g_approx}.\\

		%(optional: multiple-step update).
		\STATE Let $t \leftarrow t+1$.
		\UNTIL $t = T$
	\end{algorithmic}
\end{algorithm}
}

The latent EBM, generator and inference model can be jointly trained using stochastic gradient descent based on Eqn.\ref{eq:A}, Eqn.\ref{eq:grad_g} and Eqn.\ref{eq:grad_I}. In practice, we use sample average to approximate the expectation. 
%\vspace{0.1mm}

\noindent {\bf Synthesis and inference sampling}. The expectations for gradient computation are based on generator density $p_\theta(x, z)$ and data density $q_\phi(x, z)$. To approximate the expectation of generator density $\E_{p_\theta(x, z)}[.]$, we perform {synthesis sampling} through $z\sim p(z)$,  $\tilde{x} \sim p_\theta(x|z)$ to get $\tilde{M}$ samples $(z_i, \tilde{x}_i)$. To approximate the expectation of data density $\E_{q_\phi(x, z)}[.]$, we perform {inference sampling} through $x \sim q_{\rm data}(x), \tilde{z} \sim q_\phi(z|x)$ to get $M$ samples $(x_i, \tilde{z}_i)$. Both $p_\theta(x|z)$ and $q_\phi(z|x)$ are assumed to be Gaussian, therefore we have: 
\begin{align}
\label{eq:sample}
\tilde{x} &= g_\theta(z) + \sigma e_1, \; e_1 \sim {\rm N}(0, I_D);  \nonumber\\
\tilde{z} &= \mu_\phi(x) + V_\phi(x)^{1/2} e_2, \; e_2 \sim {\rm N}(0, I_d),
\end{align}
where $g_\theta(z)$ is the top-down deconvolutional network for the generator model (see Sec \ref{sec:generator}), and $\mu_\phi(x)$ and $V_\phi(x)$ are bottom-up convolutional networks for the mean vector and the diagonal variance-covariance matrix of the inference model (see Sec \ref{sec:inference}). We follow the common practice~\cite{goodfellow2014generative} to have $\tilde{x}$ directly from generator network, i.e., $\tilde{x} \approx g_\theta(z)$. Note that the synthesis sample $(z, \tilde{x})$ and the inference sample $(x, \tilde{z})$ are functions of the generator parameter $\theta$ and the inference parameter $\phi$ respectively which ensure gradient back-propagation. 

%therefore the synthesis sample $(z, \tilde{x})$ and the inference sample $(x, \tilde{z})$ are functions of the generator network parameter $\theta$ and the inference network parameter $\phi$ respectively, and the corresponding gradient signals can be back-propagated.

%\vspace{0.1mm}

\noindent {\bf Model learning}. The obtained synthesis samples and inference samples can be used to approximate the expectations in model learning. Specifically, for latent EBM learning, the gradient in Eqn.\ref{eq:A} can be approximated by: 
\begin{align} 
 \frac{d}{d \alpha} L_{\rm E}(\alpha) &\approx -\frac{1}{M}\sum_{i=1}^M \left[   \frac{\partial}{\partial \alpha} f_\alpha(x_i, \tilde{z}_i) \right] \nonumber \\
  &+\frac{1}{\tilde{M}} \sum_{i=1}^{\tilde{M}} \left[   \frac{\partial}{\partial \alpha} f_\alpha(\tilde{x}_i, z_i) \right]. \label{eq:A_approx}
\end{align} 
For inference model, the gradient in Eqn.\ref{eq:grad_I} can be approximated by: 
\begin{align}
\label{eq:grad_I_approx}
\frac{d}{d \phi} L_{\rm I}(\phi) &\approx \frac{\partial}{\partial \phi}\frac{1}{M}\sum_{i=1}^M\left[ \log q_{\phi}(x_i, \tilde{z}_i) - \log p_\theta(x_i, \tilde{z}_i)\right] \nonumber \\
&- \frac{\partial}{\partial \phi}\frac{1}{M}\sum_{i=1}^M\left[ \log q_{\phi}(x_i, \tilde{z}_i) - f_\alpha(x_i, \tilde{z}_i)\right].
\end{align}
For generator model, the gradient in Eqn.\ref{eq:grad_g} can be approximated by: 
\begin{align}
\label{eq:grad_g_approx}
 \frac{d}{d \theta} L_{\rm G}(\theta) &\approx - \frac{1}{M}\sum_{i=1}^M\left[ \frac{\partial}{\partial \theta}\log p_\theta(x_i|\tilde{z}_i) \right] \nonumber \\
 &- \frac{\partial}{\partial \theta}\frac{1}{\tilde{M}} \sum_{i=1}^{\tilde{M}}\left[ f_\alpha(\tilde{x}_i, z_i) \right].
\end{align}
Notice that the gradients in Eqn.\ref{eq:grad_I_approx} and Eqn.\ref{eq:grad_g_approx} on synthesis samples $(z_i, \tilde{x}_i)$ and inference samples $(x_i, \tilde{z}_i)$ can be easily back-propagated using Eqn.\ref{eq:sample}. The detailed training procedure is presented in Algorithm~\ref{alg:vae_latent}.
%The joint training algorithm iterates the above three steps by learning latent EBM, generator model and inference model , and can be efficient. The detailed training is presented in Algorithm~\ref{alg:vae_latent}.

\section{Experiments}
In this section, we evaluate the proposed model on four tasks: image generation, test image reconstruction, out-of-distribution generalization and anomaly detection. The learning of inference model is based on Eqn.\ref{eq:inf}, and we also tested the alternative way to train infernce model using Eqn.\ref{eq:inf1} for generation and reconstruction. We mainly consider 4 datasets which includes CIFAR-10, CelebA~\cite{liu2015faceattributes}, Large-scale Scene Understand (LSUN) dataset~\cite{yu2015lsun} and MNIST. We will describe the datasets in more detail in the following relevant subsections. All the training image datasets are resized and scaled to $[-1,1]$ with no further pre-processing. All network parameters are initialized with zero-mean gaussian with standard deviation 0.02 and optimized using Adam~\cite{kingma2014adam}. We adopt the similar deconvolutional network structure as in~\cite{radford2015unsupervised} for generator model and the ``mirror" convolutional structure for inference model. Both structures involve batch normalization~\cite{ioffe2015batch}. For joint energy model $\pi_\alpha(x, z)$, we use multiple layers of convolution to transform the observation $x$ and the latent factor $z$, then concatenate them at the higher layer which shares similarity as in ~\cite{li2017alice}. Spectral normalization is used as suggested in~\cite{miyato2018spectral}. We refer to our implementation \footnote{\texttt{https://hthth0801.github.io/jointLearning/}} for details.

\subsection{Image generation} 
\begin{figure*}[h]
	\begin{center}
	  \includegraphics[width=0.3\linewidth]{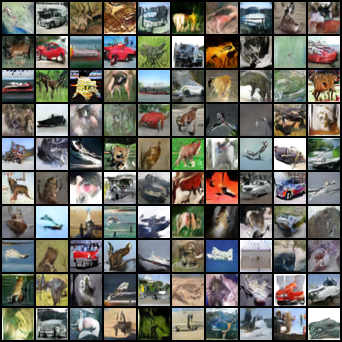}
	  \includegraphics[width=0.3\linewidth]{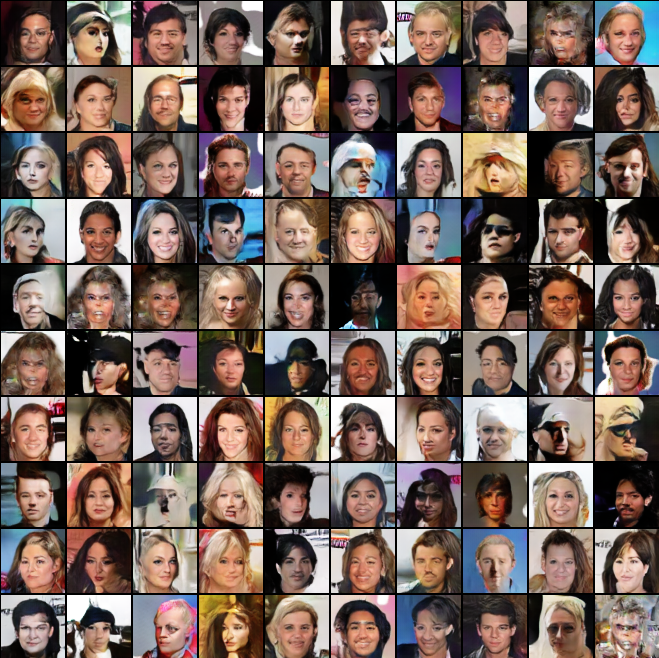}
	  \includegraphics[width=0.3\linewidth]{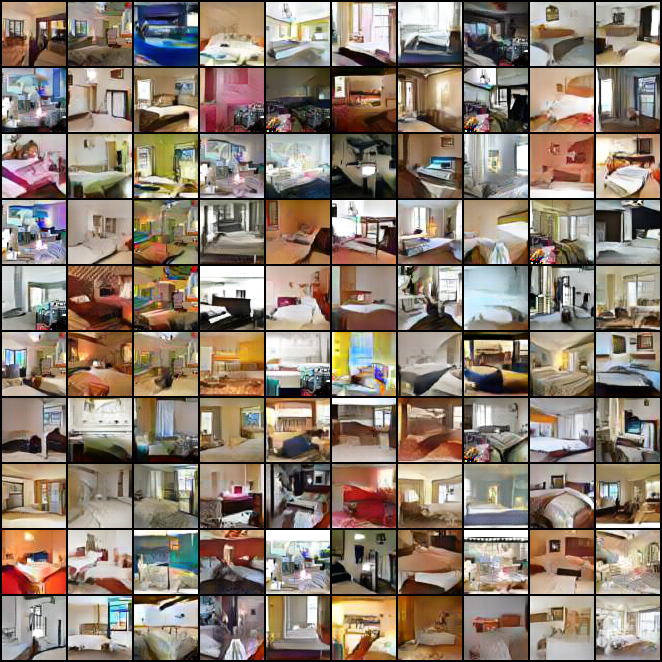}
	  \end{center}
\caption{\small Generated samples. Left: cifar10 generation. Middle: CelebA generation. Right: LSUN bedroom generation.}
\label{fig:generation}
\end{figure*}

In this experiment, we evaluate the visual quality of the generated samples. The well-learned generator network $p_\theta$ could generate samples that are realistic and share visual similarities as the training data. We mainly consider three common-used datasets including CIFAR-10, CelebA~\cite{liu2015faceattributes} and LSUN~\cite{yu2015lsun} for generation and reconstruction evaluation. The CIFAR-10 contains 60,000 color images of size $32\times 32$ of which 50,000 images are for training and 10,000 are for testing. For CelebA dataset, we resize them to be $64 \times 64$ and randomly select 10,000 images of which 9,000 are for training and 1,000 are for testing. For LSUN dataset, we select the \textit{bedroom} category which contains roughly 3 million images and resize them to $64\times 64$. We separate 10,000 images for testing and use the rest for training. The qualitative results are shown in Figure~\ref{fig:generation}. 

We further evaluate our model quantitatively by using Frechet Inception Distance (FID)~\cite{lucic2017gans} in Table~\ref{tab:fid}. We compare with baseline models including VAE~\cite{kingma2013auto}, DCGAN~\cite{radford2015unsupervised}, WGAN~\cite{arjovsky2017wasserstein}, CoopNet~\cite{xie2016cooperative}, ALICE~\cite{li2017alice}, SVAE~\cite{chen2018symmetric} and SNGAN~\cite{miyato2018spectral}. The FID scores are from the relevant papers and for missing evaluations, we re-evaluate them by utilizing their released codes or re-implement using the similar structures and optimal parameters as indicated in their papers. From Table~\ref{tab:fid}, our model achieves competitive generation performance compared to listed baseline models. Further compared to \cite{han2019divergence} which has 7.23 Inception Score (IS) on CIFAR10 and 31.9 FID on CelebA, our model has 7.17 IS and 24.7 FID respectively. It can be shown that our joint training can greatly improve the synthesis quality compared to VAE alone. Note that the SNGAN~\cite{miyato2018spectral} get better generation on CIFAR-10 which has relatively small resolution, while on other datasets that have relatively high resolution and diverse patterns, our model obtains more favorable results and has more stable training. 

\begin{table*}[h!]
  \begin{center} {\small
    \begin{tabular}{|c|c|c|c|c|c|c|c|c|c|}
    \hline Model & VAE & DCGAN & WGAN & CoopNet & ALICE & SVAE & SNGAN &  Ours(+) & Ours \\
    \hline CIFAR-10  & 	109.5  & 37.7 & 40.2 &33.6  & 48.6 &43.5 &\bf{29.3} &  33.3 & 30.1\\
    \hline CelebA  & 99.09 & 38.4 &36.4 &56.6  & 46.1 &40.7 &50.4  &  29.5 & \bf{24.7} \\
    \hline LSUN  & 175.2 & 70.4 & 67.7& 35.4 & 72 & - &67.8 &  31.4 & \bf{27.3}\\
    \hline
  \end{tabular} }
 \end{center}
 \caption{\small Sample quality evaluation using FID scores on various datasets. Ours(+) denotes our proposed method with inference  model trained using Eqn.\ref{eq:inf1}. }
  \label{tab:fid}
\end{table*}

\subsection{Testing image reconstruction}
In this experiment, we evaluate the accuracy of the learned inference model by testing image reconstruction. The well trained inference model should not only help to learn the latent EBM model but also learn to match the true posterior $p_\theta(z|x)$ of the generator model. Therefore, in practice, the well-learned inference model can be balanced to render both realistic generation as we show in previous section as well as faithful reconstruction on testing images. 

\begin{figure}[h]
	\begin{center}
	  \includegraphics[width=.48\linewidth]{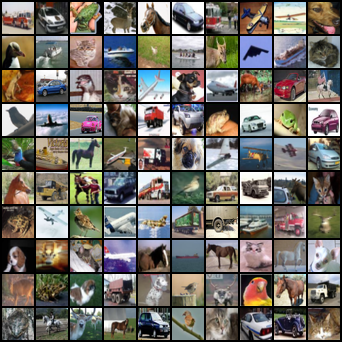}
	  \includegraphics[width=0.48\linewidth]{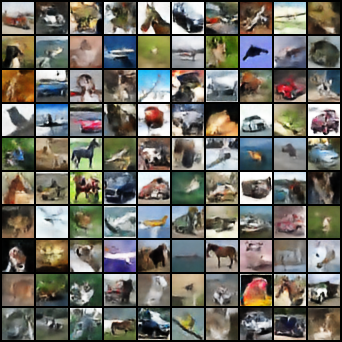}
	 \\
	 \vspace{1mm}
	   \includegraphics[width=0.48\linewidth]{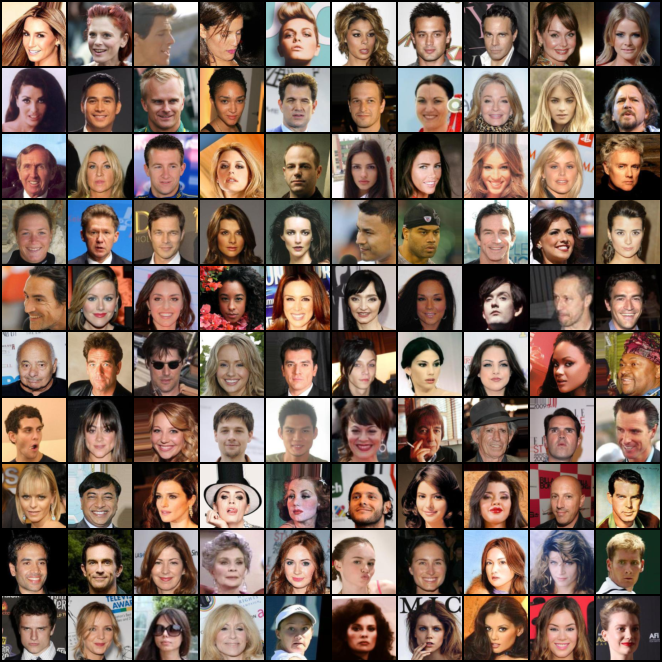}
	    \includegraphics[width=0.48\linewidth]{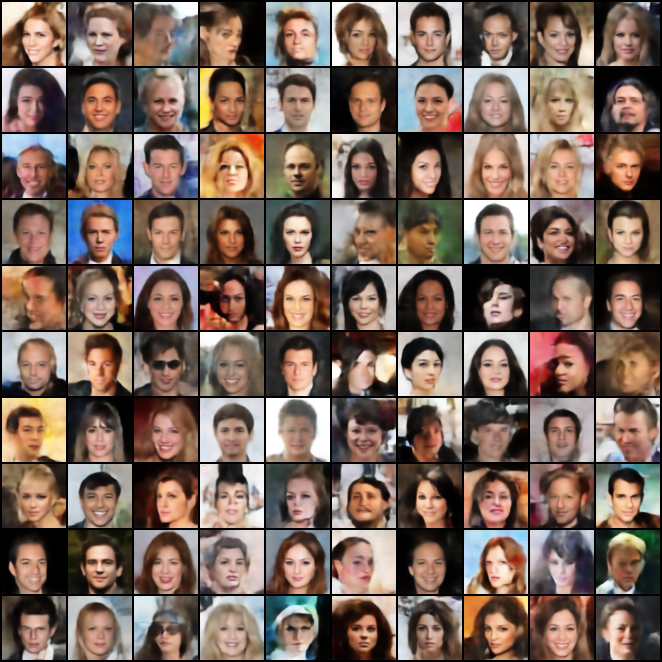}\\
	  
	  \end{center}
\caption{\small Test image reconstruction. Top: cifar10. Bottom: CelebA. Left: test images. Right: reconstructed images. }
\label{fig:test_reconstruction}
\end{figure}

We evaluate the model on hold-out testing set of CIFAR-10, CelebA and LSUN-bedroom. Specifically, we use its own 10,000 testing images for CIFAR-10, 1,000 and 10,000 hold-out testing images for CelebA and LSUN-bedroom. The testing images and the corresponding reconstructions are shown in Figure~\ref{fig:test_reconstruction}. We also quantitatively compare with baseline models (ALI~\cite{dumoulin2016adversarially}, ALICE~\cite{li2017alice}, SVAE~\cite{chen2018symmetric}) using Rooted Mean Square Error (RMSE). Note that for this experiment, we only compare with the relevant baseline models that contain joint discriminator on $(x, z)$ and could achieve the decent generation quality.  Besides, we do not consider the GANs and their variants because they have no inference model involved and are infeasible for image reconstruction. 
Table~\ref{tab:mse} shows the results. VAE is naturally integrated into our probabilistic model for joint learning. However, using VAE alone can be extremely ineffective on complex dataset.  Our model instead achieves both the high generation quality as well as the  accurate reconstruction. 

%Though it could get good testing reconstruction, it fails to generate realistic samples as shown in Table~\ref{tab:fid}.

\begin{table}[h!]
  \begin{center} {\small
    \begin{tabular}{|c|c|c|c|}
    \hline Model & CIFAR-10  & CelebA & LSUN-bedroom \\
    \hline VAE & 0.192 & 0.197 & 0.164\\
    \hline ALI & 0.558 & 0.720 & - \\
    \hline ALICE & 0.185& 0.214 & 0.181\\
    \hline SVAE & 0.258 & 0.209 & -\\
    %\hline Triangle & 0.167 &0.173 &-\\
    \hline Ours(+) & 0.184& 0.208& \bf{0.169}\\
    \hline Ours & \bf{0.177} & \bf{0.190}& \bf{0.169}\\
    %\hline
    %\hline VAE & 0.159? & 0.197 & 0.164\\
    \hline
  \end{tabular} }
 \end{center}
 \caption{Testing image reconstruction evaluation using RMSE. Ours(+) denotes our proposed method with inference  model trained using Eqn.\ref{eq:inf1}. }
%VAE results are listed for reference. Noted that VAE is naturally integrated into our model to help joint learning, however, it fails to learn the realistic generator on complex datasets.}
 \label{tab:mse}
\end{table}
\begin{figure*}[th]
	\begin{center}
	  \includegraphics[width=0.191\linewidth]{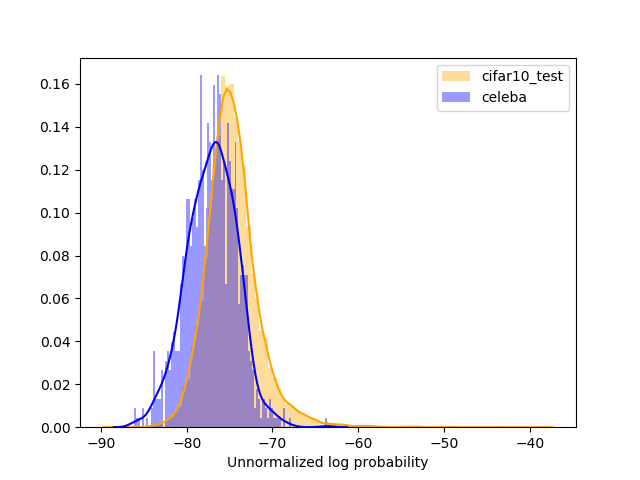}
	  \includegraphics[width=0.191\linewidth]{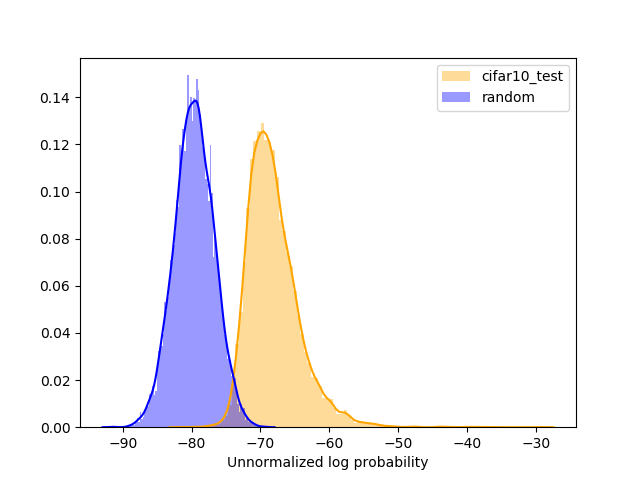}
	  \includegraphics[width=0.191\linewidth]{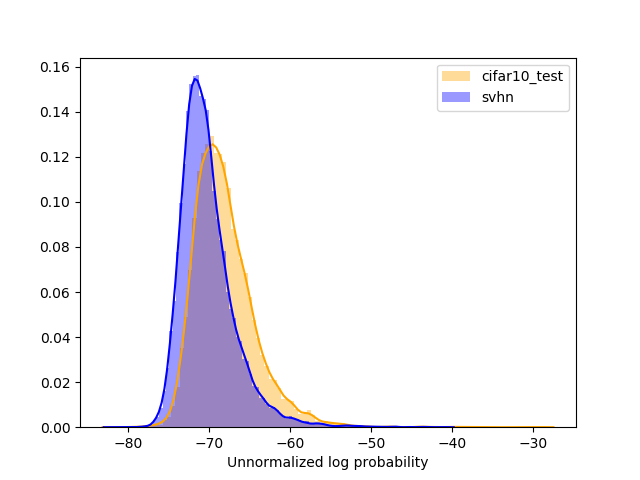}
	  \includegraphics[width=0.191\linewidth]{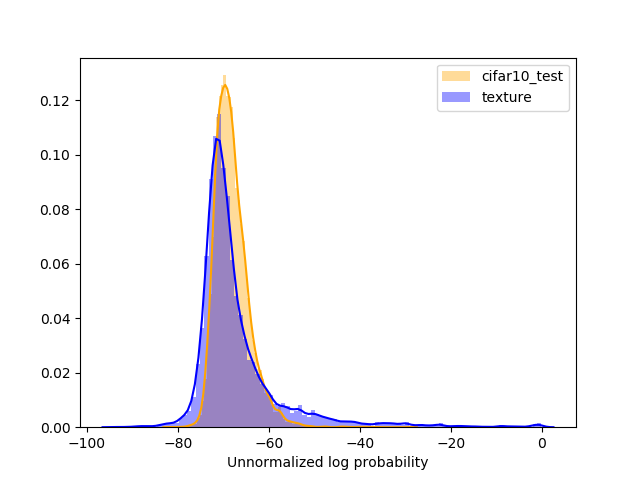}
	  \includegraphics[width=0.191\linewidth]{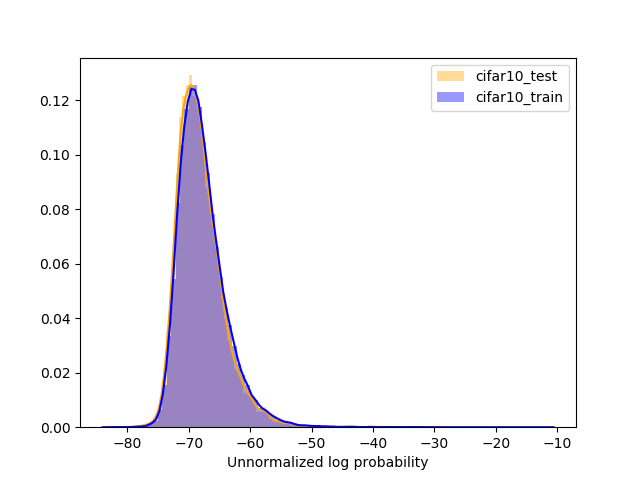}
	  \end{center}
\caption{\small Histogram of log likelihood (unnormalized) for various datasets. We provide the histogram comparison between CIFAR-10 test set and CelebA, Uniform Random, SVHN, Texture and CIFAR-10 train set respectively.}
\label{fig:ood_his}
\end{figure*}

\subsection{Out-of-distribution generalization}
In this experiment, we evaluate the out-of-distribution (OOD) detection using the learned latent EBM $\pi_\alpha(x, z)$. If the energy model is well-learned, then the training image, together with its inferred latent factor, should forms the local energy minima. Unseen images from other distribution other than training ones should be assigned to relatively higher energies. This is closely related to the model of associative memory as observed by Hopfield~\cite{hopfield1982neural}.

\begin{figure}
%\captionsetup[subfigure]{labelformat=simple}
%\setlength{\abovecaptionskip}{-4pt}
%\setlength{\belowcaptionskip}{-10pt}
\centering
\subfigure{
%\label{fig:a our method}
\begin{minipage}[b]{0.14\linewidth}
\includegraphics[width=1\textwidth]{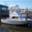}
\caption*{\small CIFAR}
\end{minipage}

}
\subfigure{
%\label{fig:b}
\begin{minipage}[b]{0.14\linewidth}
\includegraphics[width=1\textwidth]{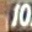}
\caption*{\small SVHN}
\end{minipage}
}
\subfigure{
%\label{fig:b}
\begin{minipage}[b]{0.14\linewidth}
\includegraphics[width=1\textwidth]{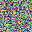}
\caption*{\small Uniform}
\end{minipage}
}
\subfigure{
%\label{fig:b}
\begin{minipage}[b]{0.14\linewidth}
\includegraphics[width=1\textwidth]{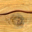}
\caption*{\small Texture}
\end{minipage}
}
\subfigure{
%\label{fig:b}
\begin{minipage}[b]{0.14\linewidth}
\includegraphics[width=1\textwidth]{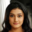}
\caption*{\small CelebA}
\end{minipage}
}
%\captionsetup{justification=flushleft}
\caption{\small Illustration of images from CIFAR-10 test, SVHN, Uniform Random, Texture and CelebA. The last four are considered to be OOD distributions.}
\label{fig:ood_examples}
\end{figure}

We learn the proposed model on CIFAR-10 training images, then utilize the learned energy model to classify the CIFAR-10 testing images from other OOD images using energy value (i.e., negative log likelihood). We use area under the ROC curve (AUROC) scores as our OOD metric following~\cite{hendrycks2016baseline} and we use Textures~\cite{cimpoi2014describing}, uniform noise, SVHN~\cite{netzer2011reading} and CelebA images as OOD distributions (Figure~\ref{fig:ood_examples} provides CIFAR10 test images and examples of OOD images). We compare with ALICE~\cite{li2017alice}, SVAE~\cite{chen2018symmetric} and the recent EBM~\cite{du2019implicit} as our baseline models. The CIFAR-10 training for ALICE, SVAE follow their optimal networks and hyperparameters and scores for EBM are taken directly from~\cite{du2019implicit}. Table~\ref{tab:auroc} shows the AUROC scores. We also provide histograms of relative likelihoods for OOD distributions in Figure~\ref{fig:ood_his} which can further verify that images from OOD distributions are assigned to relatively low log likelihood (i.e., high energy) compared to the training distribution. Our latent EBM could be learned to assign low energy for training distribution and high energy for data that comes from OOD distributions.

\begin{table}[h!]
  \begin{center} {\small
    \begin{tabular}{|c|c|c|c|c|}
    \hline Model & SVHN  & Uniform & Texture & CelebA \\
    \hline EBM & 0.63 & 1.0 & 0.48 & - \\
    \hline ALICE & 0.29& 0.0 & 0.40 & 0.48\\
    \hline SVAE & 0.42 & 0.29 & 0.5 & 0.52\\
    \hline Ours & \bf{0.68}& \bf{1.0}& \bf{0.56} & \bf{0.56}\\
    \hline
  \end{tabular}  }
 \end{center}
 \caption{\small AUROC scores of OOD classification on various images datasets. All models are learned on CIFAR-10 train set.}
 \label{tab:auroc}
\end{table}
\subsection{Anomaly detection}
In this experiment, we take a closer and more general view of the learned latent EBM with applications to anomaly detection. Unsupervised anomaly detection is one of the most important problems in machine learning and offers great potentials in many areas including cyber-security, medical analysis and surveillance etc. It is similar to the out-of-distribution detection discussed before, but can be more challenging in practice because the anomaly data may come from the distribution that is similar to and not entirely apart from the training distribution. We evaluate our model on MNIST benchmark dataset. 
\medskip

\noindent {\bf MNIST} The dataset contains 60,000 gray-scale images of size $28 \times 28$ depicting handwritten digits. Following the same experiment setting as~\cite{kumar2019maximum,zenati2018efficient}, we make each digit class an anomaly and consider the remaining 9 digits as normal examples. Our model is trained with only normal data and tested with both normal and anomalous data. We use energy function as our decision function and compare with the BiGAN-based anomaly detection model~\cite{zenati2018efficient}, the recent MEG~\cite{kumar2019maximum} and the VAE model using area under the precision-recall curve (AUPRC) as in~\cite{zenati2018efficient}. Table~\ref{tab:auprc} shows the results. 

\begin{table}[h!]
  \begin{center}
  \resizebox{1.0\columnwidth}{!}{
    \begin{tabular}{|c|c|c|c|c|}
    \hline Holdout & VAE  & MEG & BiGAN-$\sigma$ & Ours \\
    \hline 1  & 0.063  & 0.281 $\pm$ 0.035 & 0.287 $\pm$ 0.023 & \bf{0.297 $\pm$ 0.033} \\
    \hline 4  & 0.337 & 0.401 $\pm$0.061 &0.443 $\pm$ 0.029 & \bf{0.723 $\pm$ 0.042} \\
    \hline 5  & 0.325 & 0.402 $\pm$ 0.062 & 0.514 $\pm$ 0.029& \bf{0.676 $\pm$ 0.041} \\
    \hline 7  & 0.148 & 0.290 $\pm$ 0.040 & 0.347 $\pm$ 0.017& \bf{0.490 $\pm$ 0.041} \\
    \hline 9  & 0.104 & 0.342 $\pm$ 0.034 & 0.307 $\pm$ 0.028& \bf{0.383 $\pm$ 0.025} \\
    \hline
  \end{tabular}
  }
 \end{center}
 \caption{\small AUPRC scores for unsupervised anomaly detection on MNIST. Numbers are taken from~\cite{kumar2019maximum} and results for our model are averaged over last 10 epochs to account for variance.}
  \label{tab:auprc}
\end{table}

\section{Conclusion} 

This paper proposes a joint training method to learn both the VAE and the latent EBM simultaneously, where the VAE serves as an actor and the latent EBM serves as a critic. The objective function is of a simple and compact form of divergence triangle that involves three KL-divergences between three joint densities on the latent vector and the image. This objective function integrates both variational learning and adversarial learning. % The learning of generator network is based on an approximately symmetrized KL-divergence, thus correcting for the over-dispersion of VAE. The learning of latent EBM avoids the MCMC in inference sampling and synthesis sampling, thanks to the inference model and the generator network. Thus VAE and latent EBM benefit each other in the joint training. 
Our experiments show that the joint training improves the synthesis quality of VAE, and it learns reasonable energy function that is capable of anomaly detection. 

Learning well-formed energy landscape remains a challenging problem, and our experience suggests that the learned energy function can be sensitive to the setting of hyper-parameters and within the training algorithm. In our further work, we shall further improve the learning of the energy function. We shall also explore joint training of models with multiple layers of latent variables in the styles of Helmholtz machine and Boltzmann machine. 

\section*{Acknowledgment}
The work is supported by DARPA XAI project N66001-17-2-4029; ARO project W911NF1810296; and ONR MURI project N00014-16-1-2007; and Extreme Science and Engineering Discovery Environment (XSEDE) grant ASC170063. 

%\section*{Appendix} 

{\small
\bibliographystyle{ieee_fullname}
\bibliography{egbib}
}

\end{document}